\pdfoutput=1
\documentclass{article}

\usepackage{microtype}
\usepackage{graphicx}
\usepackage{subfigure}
\usepackage{booktabs} % for professional tables
\usepackage{natbib}

\usepackage{algorithm}
\usepackage{algorithmic}

\usepackage{subfigure} 
\usepackage{amsmath} 
\usepackage{amssymb} 
\usepackage{bm}
\usepackage{calrsfs}
\usepackage{multirow}
\usepackage{soul}

\usepackage{hyperref}

\usepackage[accepted]{icml2020}

\icmltitlerunning{CyCNN: A Rotation Invariant CNN}

\begin{document}

\twocolumn[
\icmltitle{CyCNN: A Rotation Invariant CNN \\ 
           using Polar Mapping and Cylindrical Convolution Layers}



\icmlsetsymbol{equal}{*}

\begin{icmlauthorlist}
\icmlauthor{Jinpyo Kim}{a}
\icmlauthor{Wookeun Jung}{a}
\icmlauthor{Hyungmo Kim}{a}
\icmlauthor{Jaejin Lee}{a}
\end{icmlauthorlist}

\icmlaffiliation{a}{Center for Manycore Programming, Seoul National University, Seoul, South Korea}
% \icmlaffiliation{b}{Seoul National University. Center for Manycore Programming.}
% \icmlaffiliation{c}{Seoul National University. Center for Manycore Programming.}
% \icmlaffiliation{d}{Seoul National University. Center for Manycore Programming.}

\icmlcorrespondingauthor{Jinpyo Kim}{jinpyo@aces.snu.ac.kr}
\icmlcorrespondingauthor{Wookeun Jung}{wookeun@aces.snu.ac.kr}
\icmlcorrespondingauthor{Hyungmo Kim}{hyungmo@aces.snu.ac.kr}
\icmlcorrespondingauthor{Jaejin Lee}{jaejin@snu.ac.kr}

\icmlkeywords{Rotational invariance, convolutional neural network,
  polar coordinates}

\vskip 0.3in
]



\printAffiliationsAndNotice{}  % leave blank if no need to mention equal contribution

\begin{abstract}
Deep Convolutional Neural Networks (CNNs) are empirically known to be invariant to moderate translation but not to rotation in image classification. This paper proposes a deep CNN model, called CyCNN, which exploits polar mapping of input images to convert rotation to translation. To deal with the cylindrical property of the polar coordinates, we replace convolution layers in conventional CNNs to cylindrical convolutional (CyConv) layers. A CyConv layer exploits the cylindrically sliding windows (CSW) mechanism that vertically extends the input-image receptive fields of boundary units in a convolutional layer. We evaluate CyCNN and conventional CNN models for classification tasks on rotated MNIST, CIFAR-10, and SVHN datasets. We show that if there is no data augmentation during training, CyCNN significantly improves classification accuracies when compared to conventional CNN models. Our implementation of CyCNN is publicly available on \url{https://github.com/mcrl/CyCNN}
\end{abstract}

\label{submission}

\section{Introduction}

Convolutional Neural Networks (CNNs) have been very successful for various computer vision tasks in the past few years. CNNs are especially well suited to tackling problems of pattern and image recognition because of their use of learned convolution filters~\cite{LBBH-IEEE-Proc-1998,  KSHI-NIPS-2012, ZEFE-ECCV-2014, SIZI-ICLR-2015, SLJS-CVPR-2015, HZRS-CVPR-2016}. Deep CNN models with some fine-tuning have achieved performance that is close to the human-level performance for image classification on various datasets~\cite{yalniz2019billionscale, touvron2019fixing}.

CNN models are empirically known to be invariant to a moderate translation of their input image even though the invariance is not explicitly encoded in them~\cite{NIPS2009_3790, SCRO-CVPR-2012, HZRS-ECCV-2014, LEVE-CoRR-2014, COWE-ICLR-2015, JSZK-NIPS-2015, COWE-ICML-2016, DFKA-ICML-2016}. This invariance becomes a beneficial property of the CNN models when we use them for image classification tasks. However, it is also known that conventional CNN models are not invariant to rotations of the input image. Such weakness leads researchers to explicitly encode rotational invariance in the model by augmenting training data or adding new structures to it. 

In this paper, we propose a rotation-invariant CNN, called \textit{CyCNN}. CyCNN is based on the following key ideas to achieve rotational invariance:
\begin{itemize}
    
    \item CyCNN converts an input image to a polar representation~\cite{SCHW-BC-1977, WECH-CGIP-1979, WIHO-CVIP-1992, BOLE-CVIU-1998}. Rotation of an image becomes translation in such a polar coordinate system.
    
    \item To deal with the cylindrical property of the polar coordinate system, CyCNN uses \textit{cylindrical convolutional} (\textsf{CyConv}) layers. A \textsf{CyConv} layer exploits \textit{cylindrically sliding windows} (CSWs) to apply its convolutional filters to its inputs. Conceptually, the CSW mechanism wraps around the input, thus transforms the input into a cylindrical shape. Then, a \textsf{CyConv} layer makes its convolutional filters to sweep the entire cylindrical input.

\end{itemize}

\begin{figure}[htbp]
    %\vskip 0.2in
    \centering
    \centerline{\includegraphics[width=0.8\linewidth, bb=100 0 650 550]{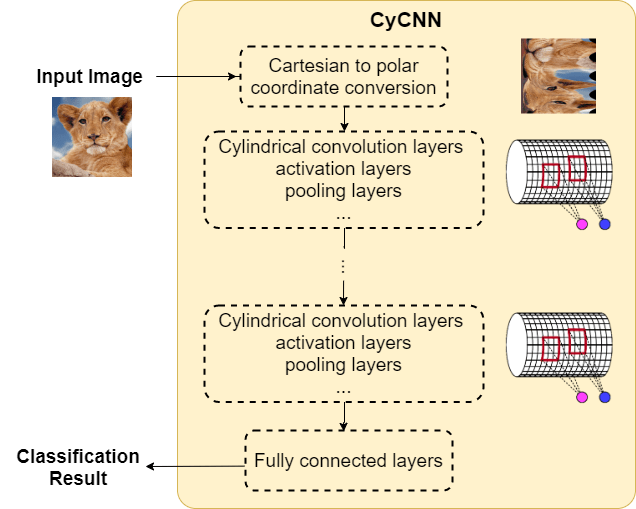}}
    \caption{The structure of CyCNN.}
    \label{fig:cycnn}
    %\vskip -0.2in
\end{figure} 

Figure~\ref{fig:cycnn} shows the structure of a CyCNN model. It first converts an input image into a polar coordinate representation. Then the converted image is processed through multiple \textsf{CyConv} layers, non-linearity layers, and pooling layers to extract feature maps of the image. Finally, fully connected layers take the resulting feature map to produce a classification result. Note that any conventional CNN can be easily transformed to CyCNN by applying polar transformation to the input, and replacing their convolutional layers with \textsf{CyConv} layers.

We evaluate some conventional CNN models and corresponding CyCNN models for classification tasks on rotated MNIST, SVHN, CIFAR-10, and CIFAR-100 datasets. We show the rotational invariance of CyCNN models by comparing their classification accuracies with those of the baseline CNN models. 

\section{Related Work}

There are several approaches to give invariance properties to CNN models. It is a common practice to augment the training set with many transformed versions of an input image to encode invariance in CNN models. Spatial transformer networks (STN)~\cite{JSZK-NIPS-2015} explicitly allows the spatial transformation of feature maps or input images to reduce pose variations in subsequent layers. The transformation is learned by the STN module in the CNN without any extra training supervision. TI-pooling layers~\cite{laptev2016ti} can efficiently handle nuisance variations in the input image caused by rotation or scaling. The layer accumulates all of the branches of activations caused by multiple transformed versions of the original input image and takes the maximum. The maximum allows the following fully connected layer to choose transformation invariant features. RIFD-CNN~\cite{CZHA-CVPR-2016} introduces two extra layers: a rotation-invariant layer and a Fisher discriminative layer. The rotation-invariant layer enforces rotation invariance on features. The Fisher discriminative layer makes the features to have small within-class scatter but large between-class separation. It uses several rotated versions of an input image for training. These approaches often result in a significant slowdown due to their computational complexities.

Another way is transforming input images or feature maps. Polar Transformer Networks (PTN)~\cite{esteves2018polar} combines ideas from the Spatial Transformer Network and canonical coordinate representations. PTN consists of a polar origin predictor, a polar transformer module, and a classifier to achieve invariance to translations and equivariance to the group of dilations/rotations. 
%The research also extends PTN model to apply on 3D object classification.
Polar Coordinate CNN (PC-CNN)~\cite{8802940} transforms input images to polar coordinates to achieve rotation-invariant feature learning. The overall structure of the model is identical to traditional CNNs except that it adopts the center loss function to learn rotation-invariant features. PC-CNN outperforms AlexNet, TI-Pooling, and Ri-CNN~\cite{7560644} on a rotated image classification test when the trained dataset is also rotation-augmented. Amorim \textit{et al.}~\cite{8489295} and Remmelzwaal \textit{et al.}~\cite{remmelzwaal2019human} analyze the effectiveness of applying the log-polar coordinate conversion to input images. Both of the approaches focus on the property that the global rotation of the original image becomes translation in the log-polar coordinate system.

Finally, by transforming convolution filters~\cite{SCRO-CVPR-2012, SOLE-ICML-2012, NIPS2014_5424, DWDA-MNRAS-2015, COWE-ICML-2016, DFKA-ICML-2016, marcos2016, worrall2017harmonic}, we can give invariance properties to CNN models. Sohn and Lee~\cite{SOLE-ICML-2012} propose a transformation-invariant restricted Boltzmann machine. It achieves the invariance of feature representation using probabilistic MAX pooling. Schmidt and Roth~\cite{SCRO-CVPR-2012} propose a general framework for incorporating transformation invariance into product models. It predicts how feature activations change as the input image is being transformed. SymNet~\cite{NIPS2014_5424} forms feature maps over arbitrary symmetry groups. It applies learnable filter-based pooling operations to achieve invariance to such symmetries. Dieleman \textit{et al.}~\cite{DWDA-MNRAS-2015} exploit rotation symmetry by rotating feature maps to solve the galaxy morphology problem. Dieleman \textit{et al.}~\cite{DFKA-ICML-2016} further extend this idea to cyclic symmetries. G-CNN~\cite{COWE-ICML-2016} shows how CNNs can be generalized to exploit larger symmetry groups including rotations and reflections. Marcos \textit{et al.}~\cite{marcos2016} propose a method for learning discriminative filters in a shallow CNN. They tie the weights of groups of filters to several rotated versions of the canonical filter of the group to extract rotation-invariant features. Harmonic Networks~\cite{worrall2017harmonic} achieve rotational invariance by replacing regular CNN filters with harmonics. These approaches of transforming convolution filters have a limitation that it is not easy to adapt the mechanism into the structure of existing models.

CyCNN does not use data augmentation nor transform convolution filters. While it applies a polar conversion to input images, it replaces the original convolutional layers with cylindrical convolutional layers to extend their receptive field. There do exist some recent studies~\cite{esteves2018polar,8802940,8489295,remmelzwaal2019human} to apply a polar conversion to input images. However, none of them considers the cylindrical property of the polar representation. 
%This motivates and lead us to the idea of cylindrical convolutional layer. 

\section{Invariances of CNNs}
There are typically six types of layers in the deep CNN architecture: an input layer, convolution layers, non-linearity (ReLU) layers, (MAX) pooling layers, fully connected layers, and an output (Softmax) layer. Pixels in the input layer or units in other layers are arranged in three dimensions: \textit{width} (denoted by $W$), \textit{height} (denoted by $H$), and \textit{channel} (denoted by $C$). Each layer before fully connected layers maps a 3D input volume to a 3D output volume. The 3D output volume is the activation of the current layer and becomes the 3D input volume to the next layer. 

\begin{figure}[ht!]
  \centering
  \includegraphics[width = 0.4\linewidth, bb=0 0 200 200]{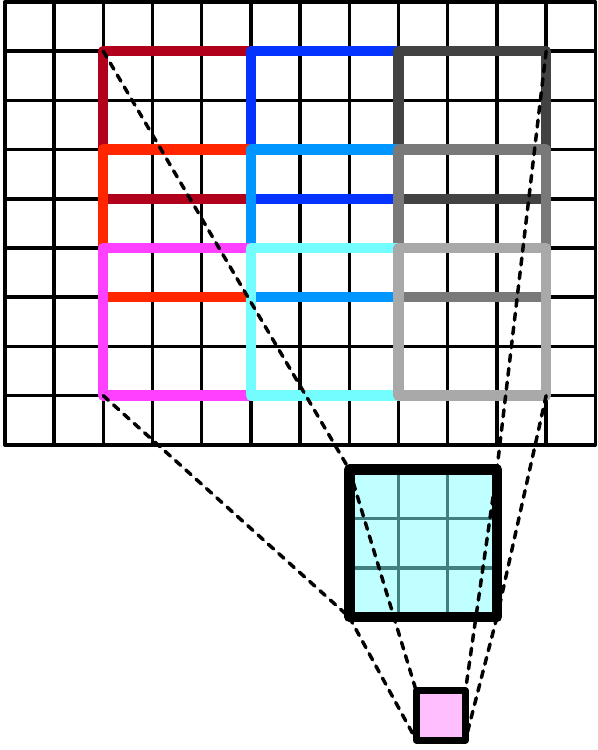}
\caption{Computing the receptive field size.
}
\label{fig:rf}
\end{figure} 

\subsection{Receptive Fields} 
The \textit{receptive field} of a unit resides in a $WH$ slice of its previous layer and is a maximal 2D region that affects the activation of the unit. Any element at the outside of the receptive field does not affect the unit~\cite{LLUZ-NIPS-2016}. The receptive field of a unit \textit{in a specific layer} (not the layer to which the unit belongs) resides in a $WH$ slice of the layer and is the maximal region that can possibly affect the unit's activation. 

Suppose that a unit in the $(i+1)$th layer has a $w_{i} \times h_{i}$ receptive field in the $i$th layer. Also, the $i$th layer has a $w_{K_{i}} \times h_{K_{i}}$ filter with strides of $s_{W_{i}}$ in width and $s_{H_{i}}$ in height. Then, the unit has a $w_{i-1} \times h_{i-1}$ receptive field in the $(i-1)$th layer in the following manner:
\[ w_{i-1} = s_{W_{i}} \cdot w_{i} + w_{K_{i}} - s_{W_{i}} \]
\[ h_{i-1} = s_{H_{i}} \cdot h_{i} + h_{K_{i}} - s_{H_{i}} \]

For example, assume that a unit has a $3 \times 3$ receptive field in its previous layer, and that the previous layer has a $3 \times 3$ filter with strides of $s_W=3$ and $s_H=2$. Then, the unit will have a $9 \times 7$ receptive field in its previous layer as shown in Figure~\ref{fig:rf}. Similarly, we can compute the input-image receptive field of a unit in any layer.

Deep CNN models increase the size of a unit's input-image receptive field by adding more convolutional and pooling layers. However, units located at the boundaries of a $WH$ slice have a much smaller input-image receptive field than units in the middle of the slice.

\subsection{Invariances}
A function $h$ is \textit{equivariant} to a transformation $g$ of input $\bm{x}$ if a corresponding transformation $g'$ of the output $h(\bm{x})$ can be found for all input $\bm{x}$, \textit{i.e.}, $h(g(\bm{x})) = g'(h(\bm{x}))$. When $g'$ is an identity function, $h$ is \textit{invariant} to $g$~\cite{SCRO-CVPR-2012, COWE-ICML-2016, DFKA-ICML-2016}. An invariant transformation is also equivariant, but not \textit{vice versa}. 

CNN models are known to be able to automatically extract invariant features to translation and small rotation/scaling using three mechanisms: \textit{local receptive fields}, \textit{parameter sharing}, and \textit{pooling}.

Pooling layers are approximately invariant to small translation, rotation, and scaling. In other words, pooling layers provide CNN models with spatial invariance to small changes in feature positions because of their filter size and selection mechanism. 

Sliding window and parameter sharing mechanisms in a convolutional layer make each unit to have a small local receptive field (the same size as its filter) that sweeps the input volume, resulting in translation equivariance of the convolutional layer. Thus, each unit in the same channel of the layer detects the same feature irrespective of the position. However, the resulting feature map is not translation invariant.

Even though a convolutional layer is not translation invariant, it builds up higher-level features by combining lower-level features. After going through the deep hierarchy of convolutional and pooling layers, a CNN model can capture more complex features. That is, each pooling layer in the hierarchy picks up more complex features because of the previous convolutional layers, and its spatial invariance to small changes in feature positions is amplified because of the previous pooling layers. 

As a result, the last pooling layer captures the highest-level features and has the strongest spatial invariance among the convolutional and pooling layers. Moreover, for a unit in the last pooling layer, the deep hierarchy makes its input-image receptive field to be the biggest. Thus, the deep hierarchy of convolutional and pooling layers enables the CNN model to integrate features over a large spatial extent in the original input image and to have moderate translation invariance.

Using a polar coordinate system and the cylindrically sliding window mechanism, CyCNN exploits and enhances such moderate translation invariance that already exists in CNN models.

\section{Achieving Rotational Invariance}
CyCNN exploits the moderate translation invariance property of CNNs to achieve rotation invariance. CyCNN converts the rotation of an input image to a translation by converting the Cartesian representation of the input image to the polar representation. Then, it applies the \textit{cylindrically sliding window} (CSW) mechanism to convolutional layers to maximize the existing translation invariance of CNNs.

\subsection{Polar Coordinate System}
% Primate visual systems provide both a wide field-of-view and a maximal resolution in the region of interest using a space-variant image mapping, such as log-polar mapping~\cite{SCHW-BC-1977, WECH-CGIP-1979, WIHO-CVIP-1992, BOLE-CVIU-1998}. While the distribution of photo-receptors in the retina is sparser in the periphery, it is denser in the fovea centrails (the central region of the retina). The image projected to the retina is in the Cartesian coordinate system. However, the primate visual system transforms it to the log-polar coordinate system. The brain transforms the coordinate system back to the Cartesian coordinates to perceive it.

% To see the properties of the polar coordinate mapping, 

We assume each pixel in an input image is a point in the Cartesian coordinate system without having any physical area occupied by itself. A point $(x, y)$ in the Cartesian coordinate system is converted to a point $(\rho, \phi)$ in the polar coordinate system as follows:

\begin{small}
\[
   \rho  =  \sqrt{x^2 + y^2} 
\]
\[
   \phi  =  \left\{ 
    \begin{array}{ll}
      \arctan(\frac{y}{x}) & \;\mbox{if}\; x > 0 \;\mbox{and}\; y \geq 0 \\
      \frac{\pi}{2} &  \;\mbox{if}\; x = 0 \;\mbox{and}\; y > 0 \\
      \pi + \arctan(\frac{y}{x}) &  \;\mbox{if}\; x < 0 \;\mbox{and}\; y \geq 0\\ 
      \pi + \arctan(\frac{y}{x}) &  \;\mbox{if}\; x < 0 \;\mbox{and}\; y < 0\\ 
      \frac{3\pi}{2} &  \;\mbox{if}\; x = 0 \;\mbox{and}\; y < 0 \\                             
      2\pi + \arctan(\frac{y}{x}) &  \;\mbox{if}\; x > 0 \;\mbox{and}\; y < 0 \\
      \mbox{undefined}  &  \;\mbox{if}\; x = 0 \;\mbox{and}\; y = 0                 
    \end{array} 
      \right.
\]
\end{small}

\begin{figure}[ht!]
  \centering
  \subfigure[]{\includegraphics[width = 0.5\linewidth, bb=0 0 550 550]{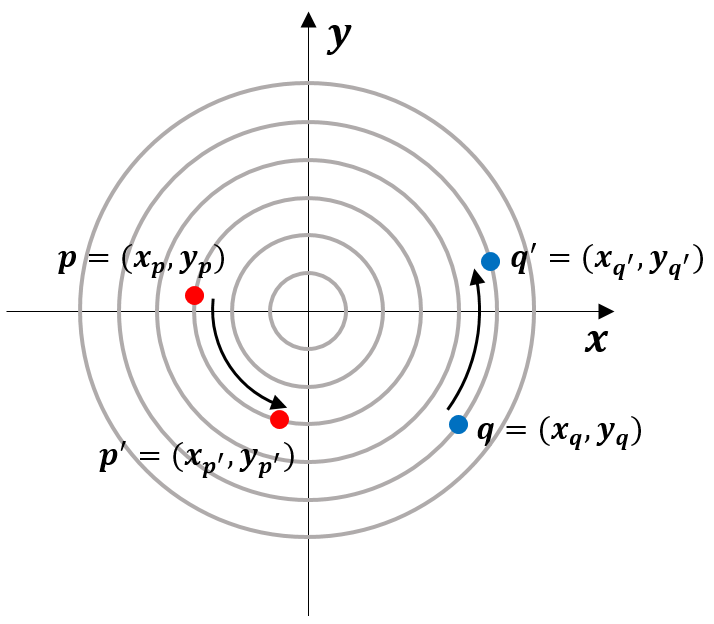}}
  ~
  \subfigure[]{\includegraphics[width = 0.45\linewidth, bb=0 0 550 550]{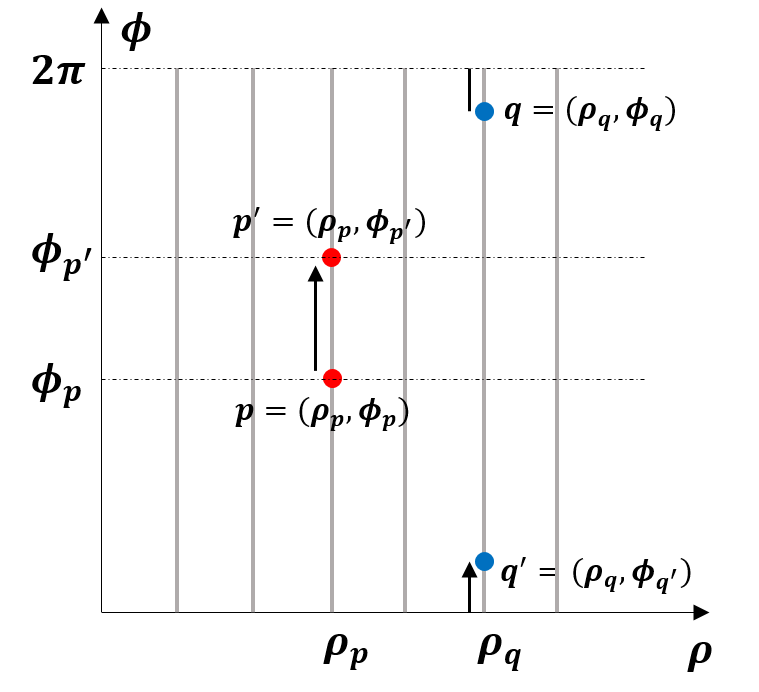}}
\caption{Converting the Cartesian coordinate system to the polar coordinate system. The concentric circles in the Cartesian coordinate system in (a) are mapped to vertical lines in the polar coordinate system in (b). 
}
\label{fig:polar-point}
\end{figure} 

Rotation in the Cartesian coordinate system become vertical translation in the polar coordinate system. For example, A point $p=(x_p, y_p)$ (colored in red) in the Cartesian coordinate system in Figure~\ref{fig:polar-point} (a) corresponds to a point $p=(\rho_p, \phi_p)$ in the polar coordinate system in (b). A point $p'$ in Figure~\ref{fig:polar-point} (a) is obtained after rotating a point $p$ around the origin $(0, 0)$ by $\phi_{p'} - \phi_p$ radians. The polar coordinate conversion maps $p$ to a point $p'=(\rho_{p'}, \phi_{p'})$ in (b). Since $\rho_{p'} = \rho_p$, the rotation becomes translation along the $\phi$ axis by $\phi_{p'} - \phi_p$ radians in (b). Note that vertical translation in polar coordinate system can go over the boundary of the image. Rotation of the point $q=(x_q, y_q)$ (colored in blue) in Figure~\ref{fig:polar-point} (a) shows the case. Since the rotation crosses the $(x>0, y=0)$ ray in the Cartesian coordinate system, its vertical translation in polar coordinate system goes over the $\phi=2\pi$ boundary as shown in Figure~\ref{fig:polar-point} (b).

The log-polar coordinate system is exactly the same as the polar coordinate system except that it takes a logarithm when calculating the distance from the origin. The calculation of $\rho$ changes into $\rho = log(\sqrt{x^2 + y^2})$. The log-polar representation of an image is inspired by the structure of the human eye and is widely used in various vision tasks~\cite{traver_bernardino_2010, 670927}. 
% However, we use (linear) polar coordinate system instead of log-polar because of the distortion of the image. 

\subsection{Input Image Conversion} 
In CyCNN, an input image is first converted to the polar or log-polar representation. Assuming that the object is placed at the center of the image, we take the center of the input image as the origin. The origin becomes the point on the bottom left corner in the polar and log-polar representations. 

\begin{figure}[htbp]
%\vskip 0.2in
\centering
  \subfigure[]{\includegraphics[width = 0.3\linewidth, bb=0 0 240 240]{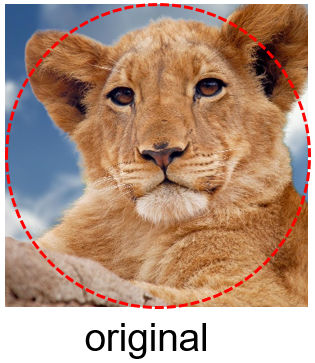}}
~
  \subfigure[]{\includegraphics[width = 0.3\linewidth, bb=0 0 240 240]{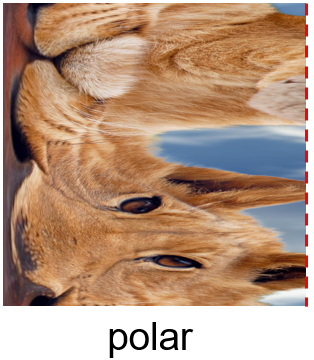}}
~
  \subfigure[]{\includegraphics[width = 0.298\linewidth, bb=0 0 240 240]{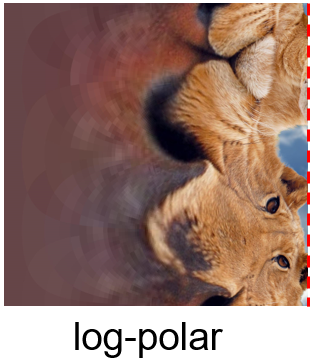}}
\vskip -0.1in
\caption{Polar and log-polar representations of an image. The red circle in (a) shows the bounding circle that indicates the maximum radius ($\rho_{max}$) of the polar coordinate. This circle is transformed to a straight line in polar and log-polar representations as shown in (b) and (c).}
\label{fig:cheetah-polar-comparison}
\end{figure}

Figure~\ref{fig:cheetah-polar-comparison} shows an example of polar and log-polar representations of an image. The polar representation already has some distortion. This is inevitable because the central and outer sections of the original image cannot preserve their area in the converted image. Furthermore, we see that the log-polar representation of the image is more distorted because the logarithm makes the central section of the original image to occupy more area. 

Note that we cannot exactly map each pixel in the original image to a pixel in the polar representation because each pixel physically occupies a space. Hence, to avoid the aliasing problem that frequently occurs in image conversion, we use the bilinear interpolation technique.

The maximum radius ($\rho_{max}$) in the polar representation is a configurable parameter in the Cartesian to polar coordinate conversion. That is, we can vary the size of \textit{bounding circle} of the original image. An example of the bounding circle is shown in Figure~\ref{fig:cheetah-polar-comparison} (a). In this paper, we set the size of the bounding circle to maximally fit in the original image as shown in Figure~\ref{fig:cheetah-polar-comparison} (a).

% Also, black regions at the right in the polar representation are \textit{outliers}, which cannot be interpolated because corresponding regions in the Cartesian representation are outside of the original image. Total area of black regions is determined by the radius of the bounding circle. By keeping the bounding circle small can reduce the amount of it. However, any feature outside the bounding circle from the original Cartesian representation will be disappeared in the polar representation. In this paper, finding the right size of the bounding circle is done by heuristics.

\begin{figure}[htbp]
\vskip 0.4in
\centering
  \subfigure[]{\includegraphics[width = 0.17\linewidth, bb=0 0 300 300]{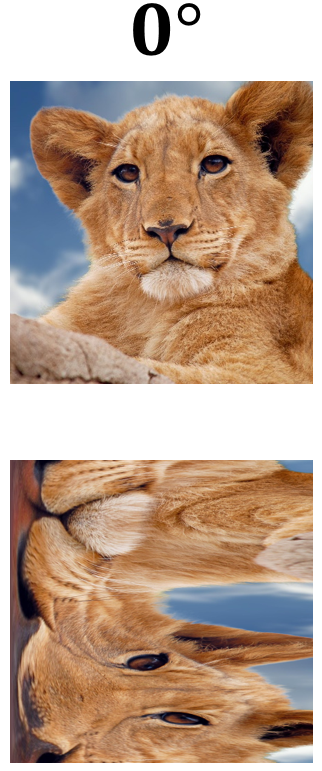}}
~
  \subfigure[]{\includegraphics[width = 0.17\linewidth, bb=0 0 300 300]{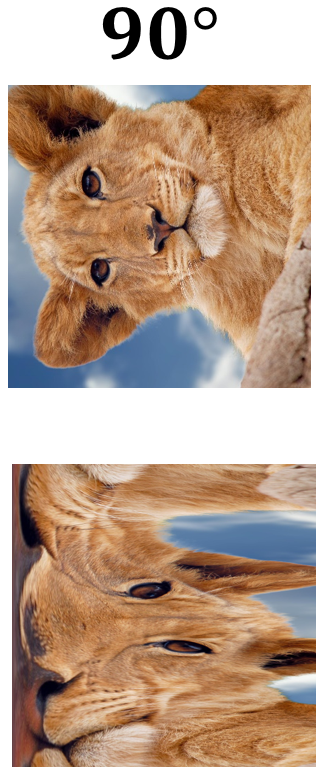}}
~
  \subfigure[]{\includegraphics[width = 0.165\linewidth, bb=0 0 300 300]{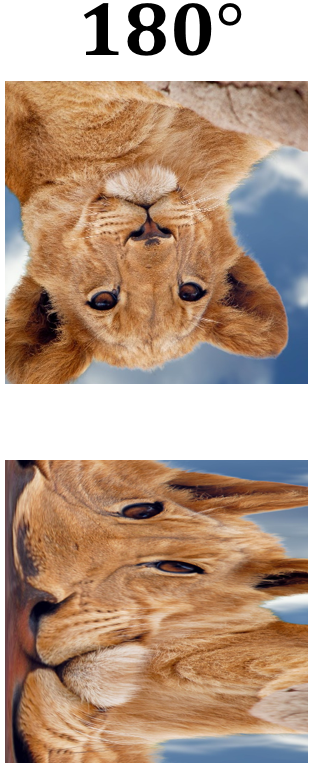}}
~
  \subfigure[]{\includegraphics[width = 0.17\linewidth, bb=0 0 300 300]{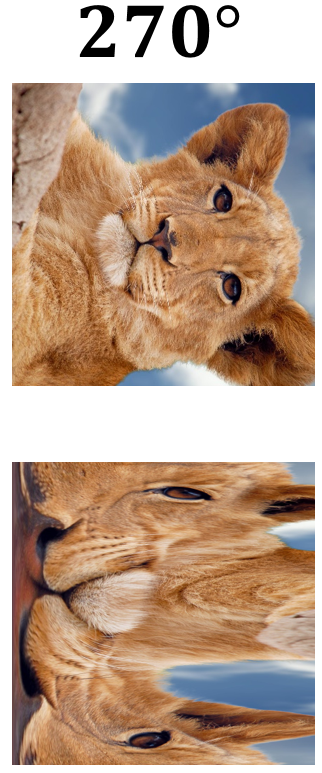}}
\vskip -0.1in
\caption{Images in the top row are generated by rotating the image in (a) by (b) 90$^{\circ}$, (c) 180$^{\circ}$, and (d) 270$^{\circ}$. Their corresponding polar coordinate representations are at the bottom row.}
\label{fig:cheetah-transform}
%\vskip -0.2in
\end{figure} 

Another example of the Cartesian to polar coordinate conversion is shown in Figure~\ref{fig:cheetah-transform}. The top row of Figure~\ref{fig:cheetah-transform} shows the results of a lion image rotated by 
90$^{\circ}$, 180$^{\circ}$, and 270$^{\circ}$ in the Cartesian representation. The bottom row shows corresponding polar representations of them. We see that rotations in the Cartesian representation become cyclic vertical translations in the polar representation.

\subsection{Cylindrically Sliding Windows}
As mentioned before, a CNN model can integrate features over a large spatial extent in the original input image. This is because the deep hierarchy of convolution and pooling layers makes the effective receptive field of each unit bigger. It also allows the CNN model to have moderate translation invariance.

\begin{figure}[ht!]
%\vskip 0.2in
\centering
  \subfigure[]{\includegraphics[width = 0.17\linewidth, bb=0 0 230 230]{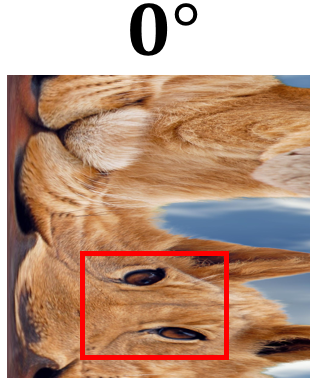}}
~
  \subfigure[]{\includegraphics[width = 0.17\linewidth, bb=0 0 230 230]{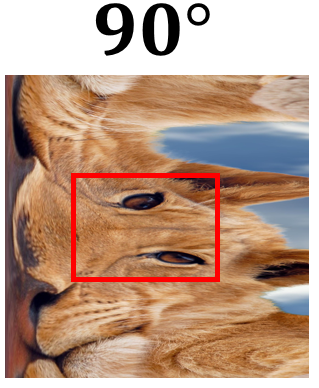}}
~
  \subfigure[]{\includegraphics[width = 0.17\linewidth, bb=0 0 230 230]{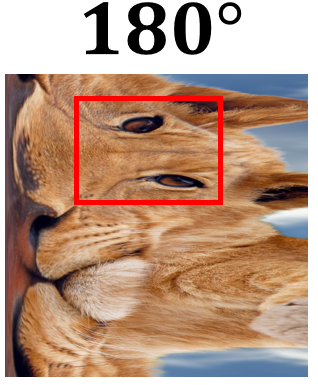}}
~
  \subfigure[]{\includegraphics[width = 0.17\linewidth, bb=0 0 230 230]{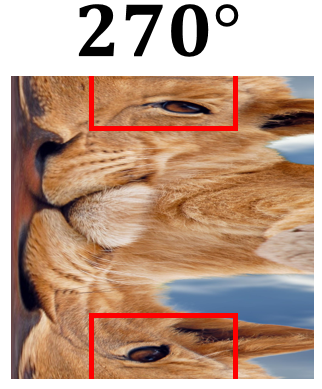}}
\vskip -0.1in
\caption{The effect of the input-image receptive field of a unit in CyCNN. (a) is the original image in the polar representation. (b), (c) and (d) are created by rotating the original image by 90$^{\circ}$, 180$^{\circ}$ and 270$^{\circ}$ respectively.}
\label{fig:erf}
%\vskip -0.2in
\end{figure} 

The input to CyCNN is an image that is in the polar representation. Consider the images (a), (b), (c), and (d) in Figure~\ref{fig:erf}. (a) is the original image represented in the polar coordinate system. (b), (c) and (d) are images created by rotating the original image by  90$^{\circ}$, 180$^{\circ}$ and 270$^{\circ}$ before represented in the polar coordinate system. The red rectangle is the input-image receptive field of a unit in some pooling layer. When a CNN model is trained with the image in (a), the unit captures and learns important features (pair of eyes in this example) in the lion's face. However, when the 270$^{\circ}$-rotated image in (d) is used as a test image, the CNN model may not recognize it as a lion because the two eyes are too far apart. Even if the receptive field captures the two eyes, the CNN model might not be able to infer them as the two eyes because their relative positions are switched. 
% Moreover, there is no way for the CNN model to recognize the two nose or mouth pieces as a whole because there is no receptive field that contains both of the pieces.

\begin{figure}[htbp]
%\vskip 0.2in
\centering
\centerline{\includegraphics[width=0.6\linewidth, bb=0 0 250 250]{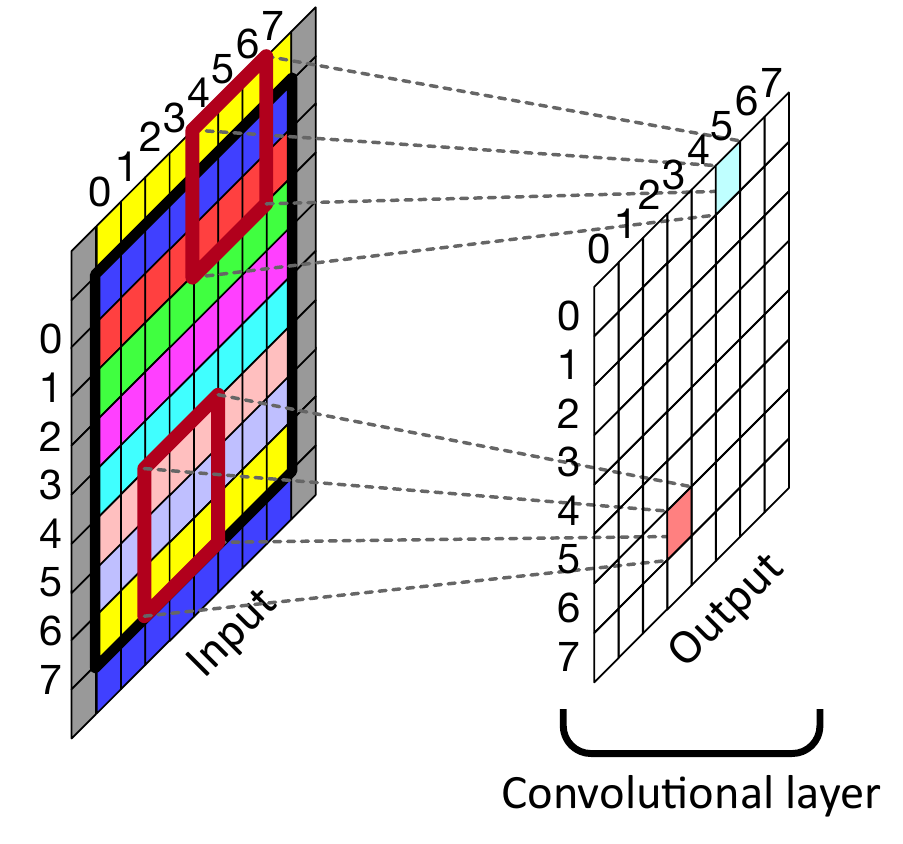}}
\caption{Cylindrically Sliding Windows (CSW) in CyCNN.}
\label{fig:sliding-window}
%\vskip -0.2in
\end{figure} 

To solve this problem, we propose \textit{cylindrically sliding windows} (CSW) for units in convolutional layers. We call such a convolutional layer a \textit{cylindrical convolutional layer} (a \textsf{CyConv} layer in short). The CSW is illustrated in Figure~\ref{fig:sliding-window}. Instead of performing zero padding at the top and bottom boundaries of the input to the convolutional layer, pixels in the first row (row 0) are copied to the boundary at the bottom, and pixels in the last row (row 7) are copied to the boundary at the top of the input. As usual, zero padding is applied to the left and right boundaries of the input. This process is the same as rolling the input vertically to make the top boundary and the bottom boundary meet together. Rolling the input in this way results in a cylinder-shaped input. The CyConv layer cyclically scans the surface of the cylindrical input with its filter.

What essentially CSW is doing is vertically extending the size of a boundary unit's receptive field in the original input image. Conceptually, CSW wraps around the input and provides it to each convolutional layer. By combining the CSW with the deep hierarchy of convolutional and pooling layers, CyCNN captures more relationships between features.

\subsection{Converting a CNN to a CyCNN}

%\begin{figure}[ht!]
%  \centering
%  \subfigure[]{\includegraphics[width = 0.405\linewidth]{figures/vgg.png}}
%  \hfill%
%  \subfigure[]{\includegraphics[width = 0.545\linewidth]{figures/cyvgg.png}}
%\caption{Structures of (a) VGG and (b) CyVGG models.}
%\label{fig:cycnn-structure}
%\end{figure} 

We can transform any CNN model into a CyCNN model easily by applying the Cartesian to polar coordinate conversion to the input image and by replacing every convolutional layer into a \textsf{CyConv} layer. Most of conventional CNNs use convolutional layers with paddings of size 1, which makes input and output feature maps have the same $WH$ size. This allows us to keep all other layers in the same configuration. 
%For example, figure~\ref{fig:cycnn-structure} (a) shows a structure of conventional VGG~\cite{SIZI-ICLR-2015} model, and Figure~\ref{fig:cycnn-structure} (b) shows a structure of corresponding CyVGG model.

Since \textsf{CyConv} layers only extend the size of the boundary unit's receptive field, a CyCNN model has exactly the same amount of learnable parameters as the corresponding original CNN model. Also, optimizations used in convolutional layers, such as the Winograd convolution algorithm~\cite{7780804}, can be applied to \textsf{CyConv} layers. The Cartesian to polar coordinate conversion takes only a small portion of overall computation. Hence, the CyCNN model requires the same amount of memory and runs at almost the same speed as the original CNN model.

\subsection{Cylindrical Winograd Convolution}
\begin{figure}[htbp]
\vskip 0.3in
\centering
\centerline{\includegraphics[width=0.8\linewidth, bb=0 0 550 550]{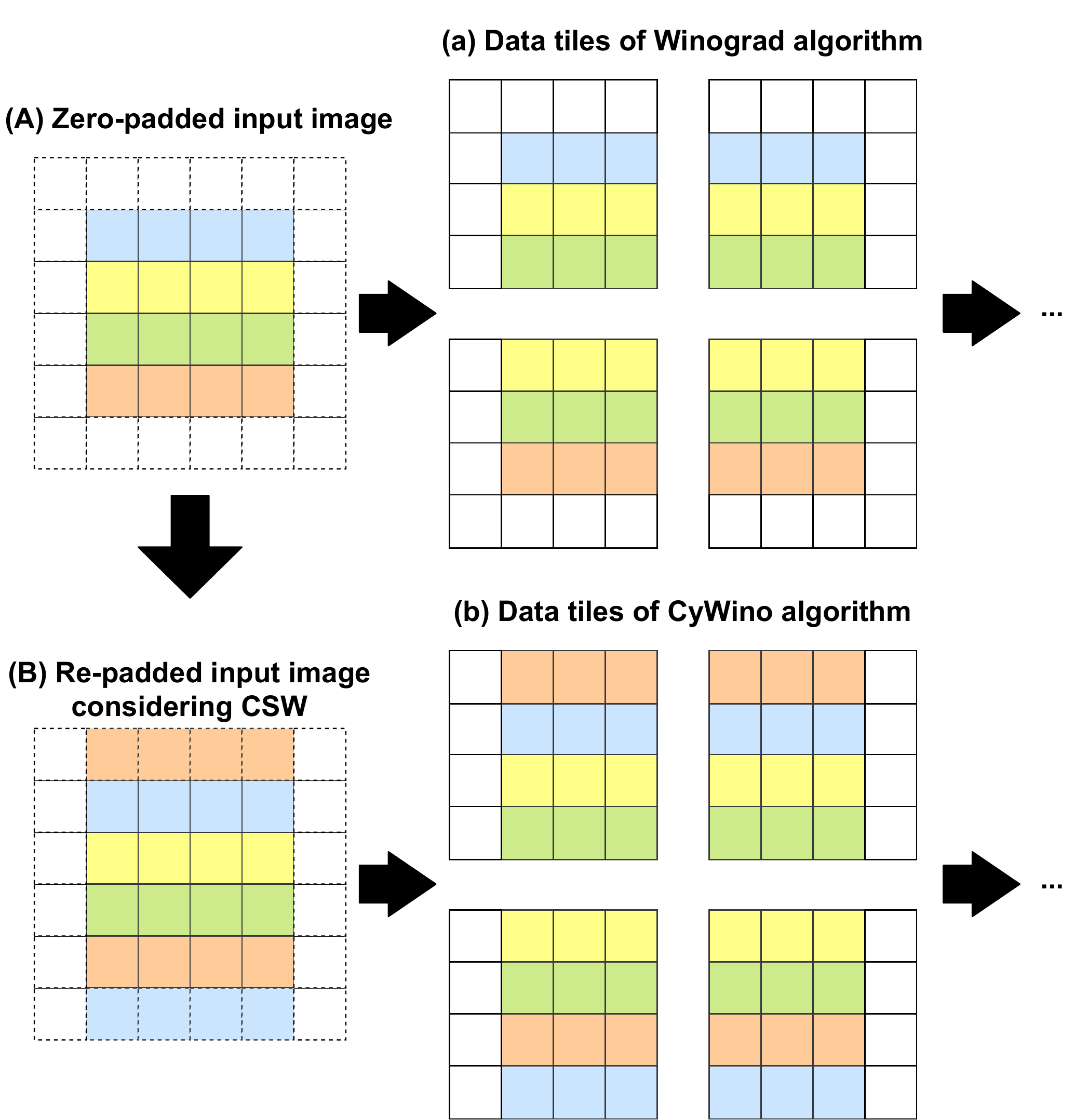}}
\caption{Data tiling phases of the Winograd algorithm and the \textsf{CyWino} algorithm.}
\label{fig:cywino}
%\vskip -0.2in
\end{figure} 

To train CyCNN in a reasonable time frame, we propose to implement a \textsf{CyConv} layer using the Winograd algorithm~\cite{7780804}. We call this layer a \textit{cylinderical Winograd convolutional} layer (a \textsf{CyWino} layer in short). 

The Winograd algorithm consists of five steps; (1) 4x4 tiles are fetched with the stride of 2 from the padded input image (i.e., the data tiling phase), (2) 3x3 filters are fetched, (3) both input tiles and filters are transformed into 4x4 Winograd tiles, (4) element-wise multiplications are performed, and (5) the results are transformed to the output features. The \textsf{CyWino} layer performs the same computation as that of the original Winograd convolution layer except for the first step.

Figure \ref{fig:cywino} describes the difference between the Winograd algorithm and the \textsf{CyWino} algorithm. In this example, a small, $4\times4$, zero-padded input image is assumed to be convolved with a $3\times3$ filter, where the padding size is 1 ((A)). In this case, we fetch four $4\times4$ tiles as shown in the figure ((a)). In the case of \textsf{CyWino}, we fill the padding considering the nature of CSW to generate a new padded input ((B)) and fetches 4x4 tiles as usual ((b)). The rest of the computations are the same as those of the original Winograd algorithm.
\section{Experiments}
In this section, we evaluate CyCNN using four image datasets: MNIST~\cite{LBBH-IEEE-Proc-1998}, Street View House Numbers (SVHN)~\cite{Netzer2011}, CIFAR-10 and CIFAR-100~\cite{Krizhevsky09learningmultiple}. 
%Note that we do not aim at achieving the state-of-the-art accuracy on these datasets. Instead, 
We are aiming at showing the effectiveness of the polar mapping and cylindrically sliding windows by comparing CyCNN models with conventional CNN models.

\subsection{CNN Models} 
We take VGG19 (with batch normalization)~\cite{SIZI-ICLR-2015} and ResNet56 ~\cite{HZRS-CVPR-2016} as our baseline models. By applying the polar transformation to the input image and replacing convolutional layers with \textsf{CyConv} layers, we obtain CyVGG19 and CyResNet56 models. Suffixes -P and -LP indicate that input images are transformed into polar and log-polar representations, respectively. 

% We also compare our CyCNN models with Ti-Pooling~\cite{laptev2016ti} model.  Core idea of this model is described in the related works section. We use author's model configuration and set number of branches to 12 and 24.

%  We set VGG19 as STN model's backend to match the scale  with other models.

\subsection{Datasets}
\textbf{MNIST}. The MNIST dataset~\cite{LBBH-IEEE-Proc-1998} is an image database of handwritten digits. It consists of a training set of 60,000 images and a test set of 10,000 images. The digits in the images have been size-normalized and centered in a fixed-size $28 \times 28$ image. To match the size of images with other datasets, every image is resized to $32 \times 32$.

\textbf{SVHN}. The Street View House Numbers~\cite{Netzer2011} (SVHN) dataset consists of over 600,000 $32 \times 32$ color images of house numbers in Google Street View. The training set consists of 75237 images, and the test set consists of 26032 images. Remaining images are extra training data that are not used in this experiment. Unlike the MNIST dataset, digits 6 and 9 are hardly distinguishable if images are rotated. Thus, we treat these two digits as the same at training/testing.

\textbf{CIFAR-10 and CIFAR-100}.  The CIFAR-10 dataset~\cite{Krizhevsky09learningmultiple} consists of 60,000 $32 \times 32$ colour images in 10 classes with 6,000 images per class. There are 50,000 training images (5,000 for each class) and 10,000 test images (1,000 images for each class) in CIFAR-10. The CIFAR-100 dataset is the same as the CIFAR-10 dataset except that it has 100 image classes. Thus, there are 500 training images and 100 test images for each class in CIFAR-100.

In every dataset, 10\% of the training set is used as the validation set. The only additional preprocessing we perform on input images is normalization.

\subsection{Implementation}
We use PyTorch~\cite{NEURIPS2019_9015} library to implement and evaluate models. We manually implement \textsf{CyConv} layers in CUDA~\cite{cuda} to train CyCNN models on GPUs. We integrate the CUDA kernels into PyTorch. We use OpenCV~\cite{opencv_library} library to implement image rotation and the polar coordinate conversion. 

%While our manually-implemented CUDA kernels for \textsf{CyConv} are functionally correct, they are too slow compared to non-cylindrical convolution primitives in cuDNN~\cite{cudnn}. For instance, when we convert convolution layers of VGG19 to \textsf{CyConv} layers, the elapsed time for training one epoch becomes $20\times$ larger.  

We manually implement the \textsf{CyWino} layer as well. When we use the \textsf{CyWino} layer, training CyVGG19~\cite{SIZI-ICLR-2015}
becomes $15\times$ faster compared to the case of using \textsf{CyConv} layer. 

When we manually implement the convolution layer using the original Winograd convolution algorithm and check its execution time, it is almost the same as that of \textsf{CyWino} layer. The \textsf{CyWino} algorithm can be integrated into highly optimized Winograd convolution implementations (\textit{e.g.} cuDNN) as well only with negligible overhead.

\subsection{Training and Testing}
We train every model using the Stochastic Gradient Descent optimizer with weight decay=$1\times10^{-5}$ and momentum=$0.9$. The cross-entropy loss is used to compare the output with the label. The learning rate is set to the initial value of 0.05, then it is reduced by half whenever the validation loss does not decrease for 5 epochs. Training completes if there is no validation accuracy improvement for 15 epochs.

In every experiment, models are tested with a rotated version of each dataset. That is, each image in the datasets is rotated by a random angle between $[0^{\circ}, 360^{\circ})$. Rotated datasets are denoted as MNIST-r, and SVHN-r, CIFAR-10-r, and CIFAR-100-r.

We train each model with four different types of training data augmentation. No augmentation (original dataset), rotation (suffixed with -r), translation (suffixed with -t), and rotation+translation (suffixed with -rt). Rotation augmentation in training is done in the same way as the test dataset. The translation augmentation randomly translates each image vertically by at most quarter of the height of the image and horizontally by at most quarter of the width of the image.

All experiments are done without any extensive hyper-parameter tuning nor fine-tuning of each model. We checked that we can obtain stable test accuracy for multiple training runs.

\subsection{Accuracy}

\begin{table}[h]
    \caption{Test accuracies on rotated datasets. Models are trained with original datasets without any data augmentation.}
    \label{tab:result1}
    \begin{center}
    \begin{scriptsize}
    %\begin{sc}
    \vskip 0.1in
    \begin{tabular}{c||cccc}
    \toprule
    Train Dataset   & MNIST     & SVHN        & CIFAR-10        & CIFAR-100\\ 
    \hline
    Test Dataset    & MNIST-r & SVHN-r    & CIFAR-10-r    & CIFAR-100-r \\
    \hline
    VGG19           & 47.20\%  & 	36.12\%  & 	32.56\%  & 	16.73\%\\
    VGG19-P         & 55.53\%  & 	43.24\%  & 	38.21\%  & 	19.96\%\\
    VGG19-LP        & 55.38\%  & 	44.76\%  & 	37.3\%  & 	18.14\%\\
    \textbf{CyVGG19-P}   & \textbf{85.49\%} & \textbf{79.77\%} & \textbf{57.58\%} & \textbf{29.76\%} \\
    \textbf{CyVGG19-LP}   & \textbf{82.90\%} & \textbf{73.91\%} & \textbf{55.94\%} & \textbf{28.32\%} \\
    ResNet56        & 44.11\%  & 	35.34\%  & 	32.05\%  & 	17.00\%\\
    ResNet56-P      & 58.95\%  & 	50.39\%  & 	38.74\%  & 	21.26\%\\
    ResNet56-LP     & 59.55\%  & 	48.95\%  & 	37.54\%  & 	20.06\%\\
    \textbf{CyResNet56-P}    & \textbf{96.71\%} & \textbf{80.25\%} & \textbf{61.27\%} & \textbf{34.10\%} \\
    \textbf{CyResNet56-LP}    & \textbf{96.84\%} & \textbf{76.71\%} & \textbf{57.08\%} & \textbf{29.15\%} \\
    \bottomrule

    \end{tabular}
    %\end{sc}
    \end{scriptsize}
    \end{center}
\end{table}

Table \ref{tab:result1} shows the classification accuracies of the models trained with original datasets without any data augmentation. Applying the polar coordinate conversion to input images increase classification accuracies in both VGG19 and ResNet56 models. It shows that applying the polar mapping to the input images is beneficial to conventional CNN models. CyCNN significantly improves classification accuracies by exploiting the cylindrical property of polar coordinates. This indicates that our approach is effective to achieve rotational invariance in CNNs. 
    
% This results show the effectiveness of both the polar coordinate conversion and the CyConv layer in rotated image classification tasks. 
% \subsection{Effects of Data Augmentation}

\begin{table}[h]
    \caption{Test accuracies on rotated datasets. Models are trained with rotation-augmented training datasets.}
    \label{tab:result2}
    \begin{center}
    \begin{scriptsize}
    %\begin{sc}
    %\vskip 0.1in
    \begin{tabular}{c||cccc}
    \toprule
    Train Dataset   & MNIST-r & SVHN-r    & CIFAR-10-r    & CIFAR-100-r\\ 
    \hline
    Test Dataset    & MNIST-r & SVHN-r    & CIFAR-10-r    & CIFAR-100-r \\
    \hline
    VGG19           & 99.61\%       & 88.70\%         & 85.61\%             & 57.87\%\\
    VGG19-P         & 99.35\%       & 88.19\%         & 75.88\%             & 44.83\%\\
    VGG19-LP        & 98.65\%       & 87.80\%         & 72.03\%             & 38.73\%\\
    \textbf{CyVGG19-P}   & \textbf{99.43\%} & \textbf{88.16\%} & \textbf{75.06\%} & \textbf{41.36\%} \\
    \textbf{CyVGG19-LP}   & \textbf{98.14\%} & \textbf{87.20\%} & \textbf{71.65\%} & \textbf{37.16\%} \\
    ResNet56        & 99.49\%       & 89.35\%         & 83.92\%             & 57.94\%\\
    ResNet56-P      & 99.41\%       & 87.87\%         & 73.16\%             & 41.99\%\\
    ResNet56-LP     & 98.71\%       & 87.86\%         & 68.05\%             & 38.33\%\\
    \textbf{CyResNet56-P}    & \textbf{99.47\%} & \textbf{87.47\%} & \textbf{71.24\%} & \textbf{41.94\%} \\
    \textbf{CyResNet56-LP}    & \textbf{98.30\%} & \textbf{87.21\%} & \textbf{67.38\%} & \textbf{37.94\%} \\
    \bottomrule

    \end{tabular}
    %\end{sc}
    \end{scriptsize}
    \end{center}
\end{table}

We also would like to see how training data augmentations affect accuracies of the models. Table~\ref{tab:result2}, Table~\ref{tab:result3}, and Table~\ref{tab:result4} contain the experimental results.

\textbf{Rotation Augmentation.} A rotational augmentation of a dataset is more beneficial to conventional CNN models because they have strength in translation but weakness in rotation. As expected, CyCNN fails to improve its classification accuracy compared to the baseline CNN models. CNN-P/LP and corresponding CyCNN models achieve almost the same classification accuracies. This implies that the reason for the loss of accuracies is the side-effect of the polar mapping: translation in the original image does not remain the same in the converted image. 

% When training images are augmented with rotation, rotational invariance is explicitly encoded in the models.

\begin{table}[h]
    \caption{Test accuracies on translated datasets. Models are trained with translation-augmented training datasets.}
    \label{tab:result3}
    \begin{center}
    \begin{scriptsize}
    %\begin{sc}
    %\vskip 0.1in
    \begin{tabular}{c||cccc}
    \toprule
    Train Dataset   & MNIST-t & SVHN-t    & CIFAR-10-t    & CIFAR-100-t\\ 
    \hline
    Test Dataset    & MNIST-r & SVHN-r    & CIFAR-10-r    & CIFAR-100-r \\
    \hline
    VGG19           & 46.98\%       & 37.92\%         & 37.15\%             & 21.21\%\\
    VGG19-P         & 52.48\%       & 46.58\%         & 45.15\%             & 30.08\%\\
    VGG19-LP        & 50.27\%       & 49.22\%         & 45.87\%             & 30.13\%\\
    \textbf{CyVGG19-P}   & \textbf{80.30\%} & \textbf{81.60\%} & \textbf{66.12\%} & \textbf{45.61\%} \\
    \textbf{CyVGG19-LP}   & \textbf{81.69\%} & \textbf{84.26\%} & \textbf{67.99\%} & \textbf{41.59\%} \\
    ResNet56        & 46.46\%       & 36.80\%         & 34.68\%             & 23.53\%\\
    ResNet56-P      & 58.29\%       & 54.62\%         & 47.80\%             & 35.64\%\\
    ResNet56-LP     & 56.71\%       & 53.49\%         & 47.29\%             & 32.03\%\\
    \textbf{CyResNet56-P}    & \textbf{94.07\%} & \textbf{84.78\%} & \textbf{68.37\%} & \textbf{50.86\%} \\
    \textbf{CyResNet56-LP}    & \textbf{96.60\%} & \textbf{88.87\%} & \textbf{73.23\%} & \textbf{46.71\%} \\
    \bottomrule

    \end{tabular}
    %\end{sc}
    \end{scriptsize}
    \end{center}
\end{table}

\textbf{Translation Augmentation.} In contrast to the rotational augmentation, a translation augmentation can benefit more to CyCNN models because they are weak to the translation of an object in the input image. That is, a feature in the original image after translation does not preserve the original shape in the polar coordinates. By comparing the results of Table~\ref{tab:result1}, Table~\ref{tab:result2} and Table~\ref{tab:result3}, we see that the classification accuracies of CyCNN models are improved significantly by the translation augmentation. MNIST dataset is an exceptional case because numbers are already positioned at the exact center of the image. Thus, the translation augmentation does not give any benefit to CyCNN for MNIST. 

% A feature in the original image after translation does not preserve the original shape in the polar coordinates.
% This is because feature in the polar-converted image does not preserve the original shape.
%CyCNN still fails to improve classification accuracies compared to baseline models trained with rotation augmented datasets.
% This downside of the polar coordinate conversion leads to the loss of translation invariance of CyCNN models. 

\begin{table}[h]
    \caption{Test accuracies on rotated and translated datasets. Models are trained with rotation+translation-augmented training datasets.}
    \label{tab:result4}
    \begin{center}
    \begin{scriptsize}
    %\begin{sc}
    %\vskip 0.1in
    \begin{tabular}{c||cccc}
    \toprule
    Train Dataset   & MNIST-rt & SVHN-rt    & CIFAR-10-rt    & CIFAR-100-rt\\ 
    \hline
    Test Dataset    & MNIST-r & SVHN-r    & CIFAR-10-r    & CIFAR-100-r \\
    \hline
    VGG19           & 99.47\%       & 93.20\%         & 83.56\%             & 58.93\%\\
    VGG19-P         & 99.29\%       & 91.50\%         & 81.90\%             & 54.68\%\\
    VGG19-LP        & 96.83\%       & 92.00\%         & 80.08\%             & 49.31\%\\
    \textbf{CyVGG19-P}   & \textbf{99.44\%} & \textbf{92.30\%} & \textbf{83.31\%} & \textbf{55.22\%} \\
    \textbf{CyVGG19-LP}   & \textbf{98.22\%} & \textbf{91.70\%} & \textbf{78.92\%} & \textbf{51.29\%} \\
    ResNet56        & 99.40\%       & 90.90\%         & 82.85\%             & 58.27\%\\
    ResNet56-P      & 99.33\%       & 88.60\%         & 79.76\%             & 53.97\%\\
    ResNet56-LP     & 97.77\%       & 87.60\%         & 79.11\%             & 52.99\%\\
    \textbf{CyResNet56-P}    & \textbf{99.38\%} & \textbf{91.60\%} & \textbf{80.24\%} & \textbf{51.25\%} \\
    \textbf{CyResNet56-LP}    & \textbf{97.41\%} & \textbf{91.10\%} & \textbf{80.30\%} & \textbf{50.78\%} \\
    \bottomrule

    \end{tabular}
    %\end{sc}
    \end{scriptsize}
    \end{center}
\end{table}

\textbf{Rotation and Translation Augmentation.} As shown in Table~\ref{tab:result4}, when both rotation and translation augmentations are applied to the training images, CyCNN models achieve competitive classification accuracies to the baseline CNN models. 

\subsection{Parameters and Training Time}

\begin{table}[htbp]    
    \caption{The number of parameters and the training time per epoch of each model on CIFAR-10 dataset. The training time is measured on a single NVIDIA Tesla V100 GPU.}
    \label{tab:params}
    \begin{center}
    \begin{scriptsize}
    \begin{tabular}{c||c|c}
    \toprule
    Model & \# Params & Training time per epoch\\
    \hline
    VGG19 & 20.6M  & 10.1 sec\\
    CyVGG19 & 20.6M & 14.8 sec\\
    ResNet56 & 0.85M & 13.9 sec\\
    CyResNet56 & 0.85M & 31.6 sec\\
    \bottomrule
    \end{tabular}
    %\end{sc}
    \end{scriptsize}
    \end{center}
\end{table}

As shown in Table~\ref{tab:params}, a CyCNN model has exactly the same number of parameters as that of its baseline CNN model. The CyCNN models run slower than baseline CNN models, especially for ResNet56. This is because our CUDA \textsf{CyWino} kernels (in CyVGG19 and CyResNet56) called by PyTorch are not fully optimized while cuDNN Winograd convolutions called by PyTorch in the baseline CNN models (VGG19 and ResNet56) are fully optimized. As mentioned earlier, we can speed up CyCNN models further by applying more optimizations to the kernels of \textsf{CyConv} layers.

\section{Conclusions}

In this paper, we propose CyCNN that exploits polar coordinate mapping and cylindrical convolutional layers to achieve rotational invariance in image classification tasks. Basically, any CNN model can be converted to a CyCNN model by applying a polar mapping to the input image and by replacing a convolutional layer with a cylindrical convolutional layer. The experimental result indicates that when the training dataset is not augmented, CyCNN has significantly better rotated-image-classification accuracy than conventional CNNs. CyCNN models still achieve competitive accuracies when both rotation and translation augmentations applied to the training images. To speedup computation in cylindrical convolutional layers, we also propose a Winograd algorithm for cylindrical convolution.

One major advantage of CyCNN is that the polar coordinate conversion and cylindrical convolution can be easily applied to any conventional CNN model without significant slowdown nor the need for more memory. We expect further studies to adapt CyCNN on various CNN models to enhance rotational invariance in their tasks.

Our implementation of CyCNN is publicly available on \url{https://github.com/mcrl/CyCNN}.

\bibliography{references}
\bibliographystyle{icml2020}

\end{document}